\documentclass[letterpaper, 10 pt, conference]{ieeeconf}  

\IEEEoverridecommandlockouts                              

\usepackage[hidelinks]{hyperref}
\usepackage{graphicx}

\usepackage{pifont}

\usepackage{tablefootnote}

\usepackage[draft]{minted}   

\makeatletter
\def\PYGdefault@reset{\let\PYGdefault@it=\relax \let\PYGdefault@bf=\relax%
    \let\PYGdefault@ul=\relax \let\PYGdefault@tc=\relax%
    \let\PYGdefault@bc=\relax \let\PYGdefault@ff=\relax}
\def\PYGdefault@tok#1{\csname PYGdefault@tok@#1\endcsname}
\def\PYGdefault@toks#1+{\ifx\relax#1\empty\else%
    \PYGdefault@tok{#1}\expandafter\PYGdefault@toks\fi}
\def\PYGdefault@do#1{\PYGdefault@bc{\PYGdefault@tc{\PYGdefault@ul{%
    \PYGdefault@it{\PYGdefault@bf{\PYGdefault@ff{#1}}}}}}}
\def\PYGdefault#1#2{\PYGdefault@reset\PYGdefault@toks#1+\relax+\PYGdefault@do{#2}}

\@namedef{PYGdefault@tok@w}{\def\PYGdefault@tc##1{\textcolor[rgb]{0.73,0.73,0.73}{##1}}}
\@namedef{PYGdefault@tok@c}{\let\PYGdefault@it=\textit\def\PYGdefault@tc##1{\textcolor[rgb]{0.24,0.48,0.48}{##1}}}
\@namedef{PYGdefault@tok@cp}{\def\PYGdefault@tc##1{\textcolor[rgb]{0.61,0.40,0.00}{##1}}}
\@namedef{PYGdefault@tok@k}{\let\PYGdefault@bf=\textbf\def\PYGdefault@tc##1{\textcolor[rgb]{0.00,0.50,0.00}{##1}}}
\@namedef{PYGdefault@tok@kp}{\def\PYGdefault@tc##1{\textcolor[rgb]{0.00,0.50,0.00}{##1}}}
\@namedef{PYGdefault@tok@kt}{\def\PYGdefault@tc##1{\textcolor[rgb]{0.69,0.00,0.25}{##1}}}
\@namedef{PYGdefault@tok@o}{\def\PYGdefault@tc##1{\textcolor[rgb]{0.40,0.40,0.40}{##1}}}
\@namedef{PYGdefault@tok@ow}{\let\PYGdefault@bf=\textbf\def\PYGdefault@tc##1{\textcolor[rgb]{0.67,0.13,1.00}{##1}}}
\@namedef{PYGdefault@tok@nb}{\def\PYGdefault@tc##1{\textcolor[rgb]{0.00,0.50,0.00}{##1}}}
\@namedef{PYGdefault@tok@nf}{\def\PYGdefault@tc##1{\textcolor[rgb]{0.00,0.00,1.00}{##1}}}
\@namedef{PYGdefault@tok@nc}{\let\PYGdefault@bf=\textbf\def\PYGdefault@tc##1{\textcolor[rgb]{0.00,0.00,1.00}{##1}}}
\@namedef{PYGdefault@tok@nn}{\let\PYGdefault@bf=\textbf\def\PYGdefault@tc##1{\textcolor[rgb]{0.00,0.00,1.00}{##1}}}
\@namedef{PYGdefault@tok@ne}{\let\PYGdefault@bf=\textbf\def\PYGdefault@tc##1{\textcolor[rgb]{0.80,0.25,0.22}{##1}}}
\@namedef{PYGdefault@tok@nv}{\def\PYGdefault@tc##1{\textcolor[rgb]{0.10,0.09,0.49}{##1}}}
\@namedef{PYGdefault@tok@no}{\def\PYGdefault@tc##1{\textcolor[rgb]{0.53,0.00,0.00}{##1}}}
\@namedef{PYGdefault@tok@nl}{\def\PYGdefault@tc##1{\textcolor[rgb]{0.46,0.46,0.00}{##1}}}
\@namedef{PYGdefault@tok@ni}{\let\PYGdefault@bf=\textbf\def\PYGdefault@tc##1{\textcolor[rgb]{0.44,0.44,0.44}{##1}}}
\@namedef{PYGdefault@tok@na}{\def\PYGdefault@tc##1{\textcolor[rgb]{0.41,0.47,0.13}{##1}}}
\@namedef{PYGdefault@tok@nt}{\let\PYGdefault@bf=\textbf\def\PYGdefault@tc##1{\textcolor[rgb]{0.00,0.50,0.00}{##1}}}
\@namedef{PYGdefault@tok@nd}{\def\PYGdefault@tc##1{\textcolor[rgb]{0.67,0.13,1.00}{##1}}}
\@namedef{PYGdefault@tok@s}{\def\PYGdefault@tc##1{\textcolor[rgb]{0.73,0.13,0.13}{##1}}}
\@namedef{PYGdefault@tok@sd}{\let\PYGdefault@it=\textit\def\PYGdefault@tc##1{\textcolor[rgb]{0.73,0.13,0.13}{##1}}}
\@namedef{PYGdefault@tok@si}{\let\PYGdefault@bf=\textbf\def\PYGdefault@tc##1{\textcolor[rgb]{0.64,0.35,0.47}{##1}}}
\@namedef{PYGdefault@tok@se}{\let\PYGdefault@bf=\textbf\def\PYGdefault@tc##1{\textcolor[rgb]{0.67,0.36,0.12}{##1}}}
\@namedef{PYGdefault@tok@sr}{\def\PYGdefault@tc##1{\textcolor[rgb]{0.64,0.35,0.47}{##1}}}
\@namedef{PYGdefault@tok@ss}{\def\PYGdefault@tc##1{\textcolor[rgb]{0.10,0.09,0.49}{##1}}}
\@namedef{PYGdefault@tok@sx}{\def\PYGdefault@tc##1{\textcolor[rgb]{0.00,0.50,0.00}{##1}}}
\@namedef{PYGdefault@tok@m}{\def\PYGdefault@tc##1{\textcolor[rgb]{0.40,0.40,0.40}{##1}}}
\@namedef{PYGdefault@tok@gh}{\let\PYGdefault@bf=\textbf\def\PYGdefault@tc##1{\textcolor[rgb]{0.00,0.00,0.50}{##1}}}
\@namedef{PYGdefault@tok@gu}{\let\PYGdefault@bf=\textbf\def\PYGdefault@tc##1{\textcolor[rgb]{0.50,0.00,0.50}{##1}}}
\@namedef{PYGdefault@tok@gd}{\def\PYGdefault@tc##1{\textcolor[rgb]{0.63,0.00,0.00}{##1}}}
\@namedef{PYGdefault@tok@gi}{\def\PYGdefault@tc##1{\textcolor[rgb]{0.00,0.52,0.00}{##1}}}
\@namedef{PYGdefault@tok@gr}{\def\PYGdefault@tc##1{\textcolor[rgb]{0.89,0.00,0.00}{##1}}}
\@namedef{PYGdefault@tok@ge}{\let\PYGdefault@it=\textit}
\@namedef{PYGdefault@tok@gs}{\let\PYGdefault@bf=\textbf}
\@namedef{PYGdefault@tok@gp}{\let\PYGdefault@bf=\textbf\def\PYGdefault@tc##1{\textcolor[rgb]{0.00,0.00,0.50}{##1}}}
\@namedef{PYGdefault@tok@go}{\def\PYGdefault@tc##1{\textcolor[rgb]{0.44,0.44,0.44}{##1}}}
\@namedef{PYGdefault@tok@gt}{\def\PYGdefault@tc##1{\textcolor[rgb]{0.00,0.27,0.87}{##1}}}
\@namedef{PYGdefault@tok@err}{\def\PYGdefault@bc##1{{\setlength{\fboxsep}{\string -\fboxrule}\fcolorbox[rgb]{1.00,0.00,0.00}{1,1,1}{\strut ##1}}}}
\@namedef{PYGdefault@tok@kc}{\let\PYGdefault@bf=\textbf\def\PYGdefault@tc##1{\textcolor[rgb]{0.00,0.50,0.00}{##1}}}
\@namedef{PYGdefault@tok@kd}{\let\PYGdefault@bf=\textbf\def\PYGdefault@tc##1{\textcolor[rgb]{0.00,0.50,0.00}{##1}}}
\@namedef{PYGdefault@tok@kn}{\let\PYGdefault@bf=\textbf\def\PYGdefault@tc##1{\textcolor[rgb]{0.00,0.50,0.00}{##1}}}
\@namedef{PYGdefault@tok@kr}{\let\PYGdefault@bf=\textbf\def\PYGdefault@tc##1{\textcolor[rgb]{0.00,0.50,0.00}{##1}}}
\@namedef{PYGdefault@tok@bp}{\def\PYGdefault@tc##1{\textcolor[rgb]{0.00,0.50,0.00}{##1}}}
\@namedef{PYGdefault@tok@fm}{\def\PYGdefault@tc##1{\textcolor[rgb]{0.00,0.00,1.00}{##1}}}
\@namedef{PYGdefault@tok@vc}{\def\PYGdefault@tc##1{\textcolor[rgb]{0.10,0.09,0.49}{##1}}}
\@namedef{PYGdefault@tok@vg}{\def\PYGdefault@tc##1{\textcolor[rgb]{0.10,0.09,0.49}{##1}}}
\@namedef{PYGdefault@tok@vi}{\def\PYGdefault@tc##1{\textcolor[rgb]{0.10,0.09,0.49}{##1}}}
\@namedef{PYGdefault@tok@vm}{\def\PYGdefault@tc##1{\textcolor[rgb]{0.10,0.09,0.49}{##1}}}
\@namedef{PYGdefault@tok@sa}{\def\PYGdefault@tc##1{\textcolor[rgb]{0.73,0.13,0.13}{##1}}}
\@namedef{PYGdefault@tok@sb}{\def\PYGdefault@tc##1{\textcolor[rgb]{0.73,0.13,0.13}{##1}}}
\@namedef{PYGdefault@tok@sc}{\def\PYGdefault@tc##1{\textcolor[rgb]{0.73,0.13,0.13}{##1}}}
\@namedef{PYGdefault@tok@dl}{\def\PYGdefault@tc##1{\textcolor[rgb]{0.73,0.13,0.13}{##1}}}
\@namedef{PYGdefault@tok@s2}{\def\PYGdefault@tc##1{\textcolor[rgb]{0.73,0.13,0.13}{##1}}}
\@namedef{PYGdefault@tok@sh}{\def\PYGdefault@tc##1{\textcolor[rgb]{0.73,0.13,0.13}{##1}}}
\@namedef{PYGdefault@tok@s1}{\def\PYGdefault@tc##1{\textcolor[rgb]{0.73,0.13,0.13}{##1}}}
\@namedef{PYGdefault@tok@mb}{\def\PYGdefault@tc##1{\textcolor[rgb]{0.40,0.40,0.40}{##1}}}
\@namedef{PYGdefault@tok@mf}{\def\PYGdefault@tc##1{\textcolor[rgb]{0.40,0.40,0.40}{##1}}}
\@namedef{PYGdefault@tok@mh}{\def\PYGdefault@tc##1{\textcolor[rgb]{0.40,0.40,0.40}{##1}}}
\@namedef{PYGdefault@tok@mi}{\def\PYGdefault@tc##1{\textcolor[rgb]{0.40,0.40,0.40}{##1}}}
\@namedef{PYGdefault@tok@il}{\def\PYGdefault@tc##1{\textcolor[rgb]{0.40,0.40,0.40}{##1}}}
\@namedef{PYGdefault@tok@mo}{\def\PYGdefault@tc##1{\textcolor[rgb]{0.40,0.40,0.40}{##1}}}
\@namedef{PYGdefault@tok@ch}{\let\PYGdefault@it=\textit\def\PYGdefault@tc##1{\textcolor[rgb]{0.24,0.48,0.48}{##1}}}
\@namedef{PYGdefault@tok@cm}{\let\PYGdefault@it=\textit\def\PYGdefault@tc##1{\textcolor[rgb]{0.24,0.48,0.48}{##1}}}
\@namedef{PYGdefault@tok@cpf}{\let\PYGdefault@it=\textit\def\PYGdefault@tc##1{\textcolor[rgb]{0.24,0.48,0.48}{##1}}}
\@namedef{PYGdefault@tok@c1}{\let\PYGdefault@it=\textit\def\PYGdefault@tc##1{\textcolor[rgb]{0.24,0.48,0.48}{##1}}}
\@namedef{PYGdefault@tok@cs}{\let\PYGdefault@it=\textit\def\PYGdefault@tc##1{\textcolor[rgb]{0.24,0.48,0.48}{##1}}}

\makeatother

\makeatletter
\def\PYG@reset{\let\PYG@it=\relax \let\PYG@bf=\relax%
    \let\PYG@ul=\relax \let\PYG@tc=\relax%
    \let\PYG@bc=\relax \let\PYG@ff=\relax}
\def\PYG@tok#1{\csname PYG@tok@#1\endcsname}
\def\PYG@toks#1+{\ifx\relax#1\empty\else%
    \PYG@tok{#1}\expandafter\PYG@toks\fi}
\def\PYG@do#1{\PYG@bc{\PYG@tc{\PYG@ul{%
    \PYG@it{\PYG@bf{\PYG@ff{#1}}}}}}}
\def\PYG#1#2{\PYG@reset\PYG@toks#1+\relax+\PYG@do{#2}}

\@namedef{PYG@tok@w}{\def\PYG@tc##1{\textcolor[rgb]{0.73,0.73,0.73}{##1}}}
\@namedef{PYG@tok@c}{\let\PYG@it=\textit\def\PYG@tc##1{\textcolor[rgb]{0.24,0.48,0.48}{##1}}}
\@namedef{PYG@tok@cp}{\def\PYG@tc##1{\textcolor[rgb]{0.61,0.40,0.00}{##1}}}
\@namedef{PYG@tok@k}{\let\PYG@bf=\textbf\def\PYG@tc##1{\textcolor[rgb]{0.00,0.50,0.00}{##1}}}
\@namedef{PYG@tok@kp}{\def\PYG@tc##1{\textcolor[rgb]{0.00,0.50,0.00}{##1}}}
\@namedef{PYG@tok@kt}{\def\PYG@tc##1{\textcolor[rgb]{0.69,0.00,0.25}{##1}}}
\@namedef{PYG@tok@o}{\def\PYG@tc##1{\textcolor[rgb]{0.40,0.40,0.40}{##1}}}
\@namedef{PYG@tok@ow}{\let\PYG@bf=\textbf\def\PYG@tc##1{\textcolor[rgb]{0.67,0.13,1.00}{##1}}}
\@namedef{PYG@tok@nb}{\def\PYG@tc##1{\textcolor[rgb]{0.00,0.50,0.00}{##1}}}
\@namedef{PYG@tok@nf}{\def\PYG@tc##1{\textcolor[rgb]{0.00,0.00,1.00}{##1}}}
\@namedef{PYG@tok@nc}{\let\PYG@bf=\textbf\def\PYG@tc##1{\textcolor[rgb]{0.00,0.00,1.00}{##1}}}
\@namedef{PYG@tok@nn}{\let\PYG@bf=\textbf\def\PYG@tc##1{\textcolor[rgb]{0.00,0.00,1.00}{##1}}}
\@namedef{PYG@tok@ne}{\let\PYG@bf=\textbf\def\PYG@tc##1{\textcolor[rgb]{0.80,0.25,0.22}{##1}}}
\@namedef{PYG@tok@nv}{\def\PYG@tc##1{\textcolor[rgb]{0.10,0.09,0.49}{##1}}}
\@namedef{PYG@tok@no}{\def\PYG@tc##1{\textcolor[rgb]{0.53,0.00,0.00}{##1}}}
\@namedef{PYG@tok@nl}{\def\PYG@tc##1{\textcolor[rgb]{0.46,0.46,0.00}{##1}}}
\@namedef{PYG@tok@ni}{\let\PYG@bf=\textbf\def\PYG@tc##1{\textcolor[rgb]{0.44,0.44,0.44}{##1}}}
\@namedef{PYG@tok@na}{\def\PYG@tc##1{\textcolor[rgb]{0.41,0.47,0.13}{##1}}}
\@namedef{PYG@tok@nt}{\let\PYG@bf=\textbf\def\PYG@tc##1{\textcolor[rgb]{0.00,0.50,0.00}{##1}}}
\@namedef{PYG@tok@nd}{\def\PYG@tc##1{\textcolor[rgb]{0.67,0.13,1.00}{##1}}}
\@namedef{PYG@tok@s}{\def\PYG@tc##1{\textcolor[rgb]{0.73,0.13,0.13}{##1}}}
\@namedef{PYG@tok@sd}{\let\PYG@it=\textit\def\PYG@tc##1{\textcolor[rgb]{0.73,0.13,0.13}{##1}}}
\@namedef{PYG@tok@si}{\let\PYG@bf=\textbf\def\PYG@tc##1{\textcolor[rgb]{0.64,0.35,0.47}{##1}}}
\@namedef{PYG@tok@se}{\let\PYG@bf=\textbf\def\PYG@tc##1{\textcolor[rgb]{0.67,0.36,0.12}{##1}}}
\@namedef{PYG@tok@sr}{\def\PYG@tc##1{\textcolor[rgb]{0.64,0.35,0.47}{##1}}}
\@namedef{PYG@tok@ss}{\def\PYG@tc##1{\textcolor[rgb]{0.10,0.09,0.49}{##1}}}
\@namedef{PYG@tok@sx}{\def\PYG@tc##1{\textcolor[rgb]{0.00,0.50,0.00}{##1}}}
\@namedef{PYG@tok@m}{\def\PYG@tc##1{\textcolor[rgb]{0.40,0.40,0.40}{##1}}}
\@namedef{PYG@tok@gh}{\let\PYG@bf=\textbf\def\PYG@tc##1{\textcolor[rgb]{0.00,0.00,0.50}{##1}}}
\@namedef{PYG@tok@gu}{\let\PYG@bf=\textbf\def\PYG@tc##1{\textcolor[rgb]{0.50,0.00,0.50}{##1}}}
\@namedef{PYG@tok@gd}{\def\PYG@tc##1{\textcolor[rgb]{0.63,0.00,0.00}{##1}}}
\@namedef{PYG@tok@gi}{\def\PYG@tc##1{\textcolor[rgb]{0.00,0.52,0.00}{##1}}}
\@namedef{PYG@tok@gr}{\def\PYG@tc##1{\textcolor[rgb]{0.89,0.00,0.00}{##1}}}
\@namedef{PYG@tok@ge}{\let\PYG@it=\textit}
\@namedef{PYG@tok@gs}{\let\PYG@bf=\textbf}
\@namedef{PYG@tok@gp}{\let\PYG@bf=\textbf\def\PYG@tc##1{\textcolor[rgb]{0.00,0.00,0.50}{##1}}}
\@namedef{PYG@tok@go}{\def\PYG@tc##1{\textcolor[rgb]{0.44,0.44,0.44}{##1}}}
\@namedef{PYG@tok@gt}{\def\PYG@tc##1{\textcolor[rgb]{0.00,0.27,0.87}{##1}}}
\@namedef{PYG@tok@err}{\def\PYG@bc##1{{\setlength{\fboxsep}{\string -\fboxrule}\fcolorbox[rgb]{1.00,0.00,0.00}{1,1,1}{\strut ##1}}}}
\@namedef{PYG@tok@kc}{\let\PYG@bf=\textbf\def\PYG@tc##1{\textcolor[rgb]{0.00,0.50,0.00}{##1}}}
\@namedef{PYG@tok@kd}{\let\PYG@bf=\textbf\def\PYG@tc##1{\textcolor[rgb]{0.00,0.50,0.00}{##1}}}
\@namedef{PYG@tok@kn}{\let\PYG@bf=\textbf\def\PYG@tc##1{\textcolor[rgb]{0.00,0.50,0.00}{##1}}}
\@namedef{PYG@tok@kr}{\let\PYG@bf=\textbf\def\PYG@tc##1{\textcolor[rgb]{0.00,0.50,0.00}{##1}}}
\@namedef{PYG@tok@bp}{\def\PYG@tc##1{\textcolor[rgb]{0.00,0.50,0.00}{##1}}}
\@namedef{PYG@tok@fm}{\def\PYG@tc##1{\textcolor[rgb]{0.00,0.00,1.00}{##1}}}
\@namedef{PYG@tok@vc}{\def\PYG@tc##1{\textcolor[rgb]{0.10,0.09,0.49}{##1}}}
\@namedef{PYG@tok@vg}{\def\PYG@tc##1{\textcolor[rgb]{0.10,0.09,0.49}{##1}}}
\@namedef{PYG@tok@vi}{\def\PYG@tc##1{\textcolor[rgb]{0.10,0.09,0.49}{##1}}}
\@namedef{PYG@tok@vm}{\def\PYG@tc##1{\textcolor[rgb]{0.10,0.09,0.49}{##1}}}
\@namedef{PYG@tok@sa}{\def\PYG@tc##1{\textcolor[rgb]{0.73,0.13,0.13}{##1}}}
\@namedef{PYG@tok@sb}{\def\PYG@tc##1{\textcolor[rgb]{0.73,0.13,0.13}{##1}}}
\@namedef{PYG@tok@sc}{\def\PYG@tc##1{\textcolor[rgb]{0.73,0.13,0.13}{##1}}}
\@namedef{PYG@tok@dl}{\def\PYG@tc##1{\textcolor[rgb]{0.73,0.13,0.13}{##1}}}
\@namedef{PYG@tok@s2}{\def\PYG@tc##1{\textcolor[rgb]{0.73,0.13,0.13}{##1}}}
\@namedef{PYG@tok@sh}{\def\PYG@tc##1{\textcolor[rgb]{0.73,0.13,0.13}{##1}}}
\@namedef{PYG@tok@s1}{\def\PYG@tc##1{\textcolor[rgb]{0.73,0.13,0.13}{##1}}}
\@namedef{PYG@tok@mb}{\def\PYG@tc##1{\textcolor[rgb]{0.40,0.40,0.40}{##1}}}
\@namedef{PYG@tok@mf}{\def\PYG@tc##1{\textcolor[rgb]{0.40,0.40,0.40}{##1}}}
\@namedef{PYG@tok@mh}{\def\PYG@tc##1{\textcolor[rgb]{0.40,0.40,0.40}{##1}}}
\@namedef{PYG@tok@mi}{\def\PYG@tc##1{\textcolor[rgb]{0.40,0.40,0.40}{##1}}}
\@namedef{PYG@tok@il}{\def\PYG@tc##1{\textcolor[rgb]{0.40,0.40,0.40}{##1}}}
\@namedef{PYG@tok@mo}{\def\PYG@tc##1{\textcolor[rgb]{0.40,0.40,0.40}{##1}}}
\@namedef{PYG@tok@ch}{\let\PYG@it=\textit\def\PYG@tc##1{\textcolor[rgb]{0.24,0.48,0.48}{##1}}}
\@namedef{PYG@tok@cm}{\let\PYG@it=\textit\def\PYG@tc##1{\textcolor[rgb]{0.24,0.48,0.48}{##1}}}
\@namedef{PYG@tok@cpf}{\let\PYG@it=\textit\def\PYG@tc##1{\textcolor[rgb]{0.24,0.48,0.48}{##1}}}
\@namedef{PYG@tok@c1}{\let\PYG@it=\textit\def\PYG@tc##1{\textcolor[rgb]{0.24,0.48,0.48}{##1}}}
\@namedef{PYG@tok@cs}{\let\PYG@it=\textit\def\PYG@tc##1{\textcolor[rgb]{0.24,0.48,0.48}{##1}}}

\def\PYGZus{\char`\_}

\def\PYGZsh{\char`\#}

\def\PYGZhy{\char`\-}
\def\PYGZsq{\char`\'}

\makeatother

\usepackage{amsmath, amsfonts}
\usepackage{caption}
\usepackage{subcaption}

\title{\LARGE \bf
  OpTaS: An Optimization-based Task Specification Library for Trajectory Optimization and Model Predictive Control
}

\author{
  Christopher E. Mower,
  Jo\~{a}o Moura,
  Nazanin Zamani Behabadi,\\
  Sethu Vijayakumar,
  Tom Vercauteren$^{*}$,
  Christos Bergeles$^{*}$
  \thanks{C.~E.~Mower, C.~Bergeles and T.~Vercauteren are with the School of Biomedical Engineering \& Imaging Sciences, King's College London, UK. J.~Moura and S.~Vijaykumar are with School of Informatics, University of Edinburgh, UK. Correspondence: \href{mailto:chris.mower@kcl.ac.uk}{christopher.mower@kcl.ac.uk}.}
  \thanks{This research  received funding from the European Union’s Horizon 2020 research and innovation program under grant agreement No. 101016985 (FAROS). Further, this work was supported by core funding from the Wellcome/EPSRC [WT203148/Z/16/Z; NS/A000049/1]. T.~ Vercauteren is supported by a Medtronic / RAEng Research Chair [RCSRF1819\textbackslash7\textbackslash34], and C.~Bergeles by an ERC Starting Grant [714562]. This work has received funding
from the European Union’s Horizon 2020 research and innovation programme under grant agreement No 101017008, Enhancing Healthcare with Assistive Robotic Mobile Manipulation (HARMONY). This work was supported by core funding from the Wellcome/EPSRC [WT203148/Z/16/Z;
NS/A000049/1].
  This research is supported by Kawada Robotics Corporation, Japan and the Alan Turing Institute, UK.  
} 
\thanks{$^{*}$C.~Bergeles and T.~Vercauteren equally contributed to the work.}
\thanks{For the purpose of open access, the authors have applied a CC BY public copyright license to any Author Accepted Manuscript version arising from this submission.}
}

\begin{document}

\maketitle
\thispagestyle{empty}
\pagestyle{empty}

\begin{abstract}
  This paper presents OpTaS, a task specification Python library for Trajectory Optimization (TO) and Model Predictive Control (MPC) in robotics. Both TO and MPC are increasingly receiving interest in optimal control and in particular handling dynamic environments. 
  While a flurry of software libraries exists to handle such problems, they either provide interfaces that are limited to a specific problem formulation (e.g. TracIK, CHOMP), or are large and statically specify the problem  in configuration files (e.g. EXOTica, eTaSL).  
  OpTaS, on the other hand, allows a user to specify custom nonlinear constrained problem formulations in a single Python script allowing the controller parameters to be modified during execution.
  The library provides interface to several open source and commercial solvers (e.g. IPOPT, SNOPT, KNITRO, SciPy) to facilitate integration with established workflows in robotics. Further benefits of OpTaS are highlighted through a thorough comparison with common libraries. 
  An additional key advantage of OpTaS is the ability to define optimal control tasks in the joint space, task space, or indeed simultaneously.
  The code for OpTaS is easily installed via \texttt{pip}, and the source code with examples can be found at \href{https://github.com/cmower/optas}{github.com/cmower/optas}.
\end{abstract}

\section{Introduction}\label{sec:intro}

\begin{figure}
  \def\figheight{2.7cm}
  \def\colwidth{0.4775\columnwidth}
  \centering
  \begin{subfigure}[b]{\colwidth}
    \centering
    \includegraphics[height=\figheight]{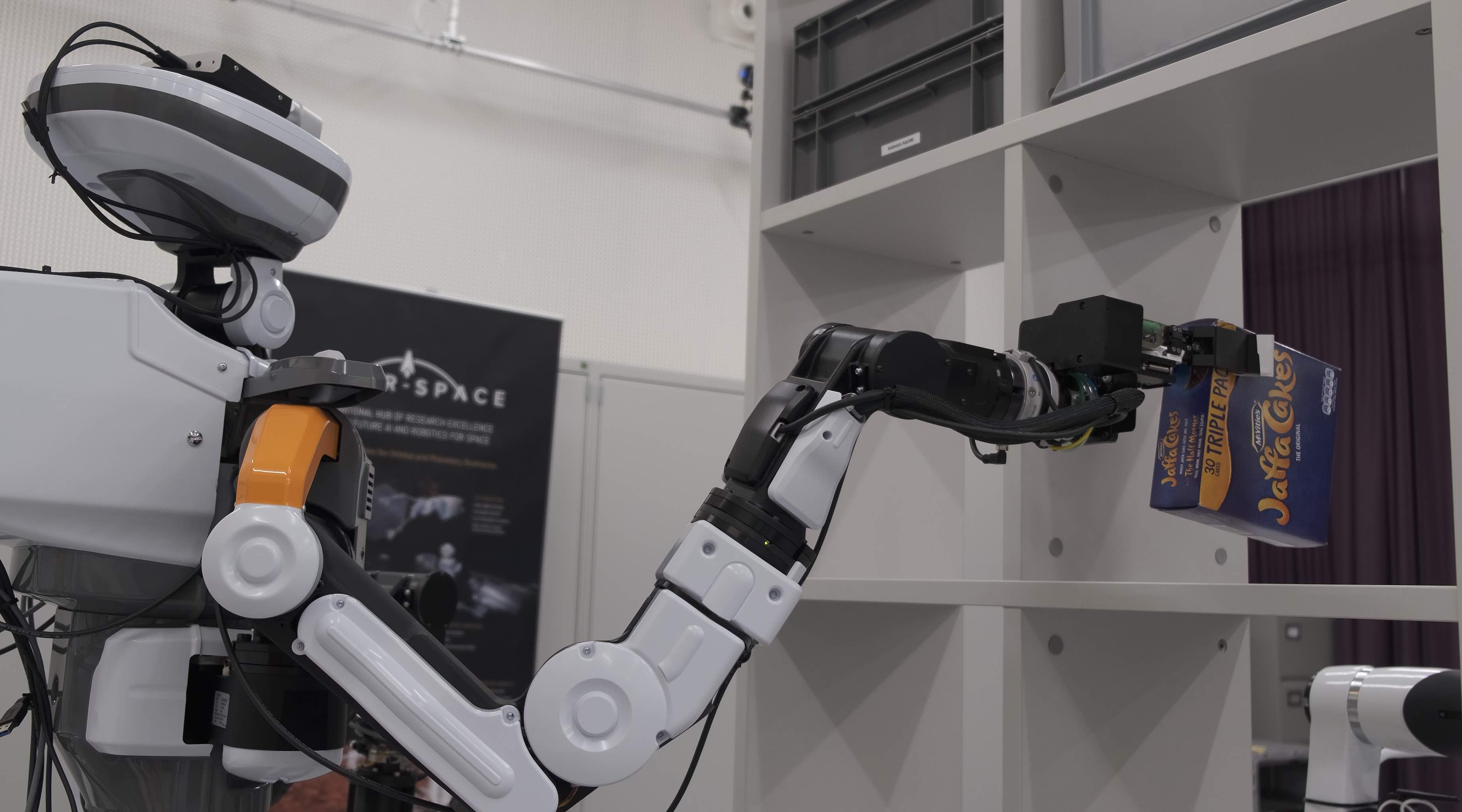}
    \caption{}
    \label{fig:shelf}
  \end{subfigure}
  ~
  \begin{subfigure}[b]{\colwidth}
    \centering
    \includegraphics[height=\figheight]{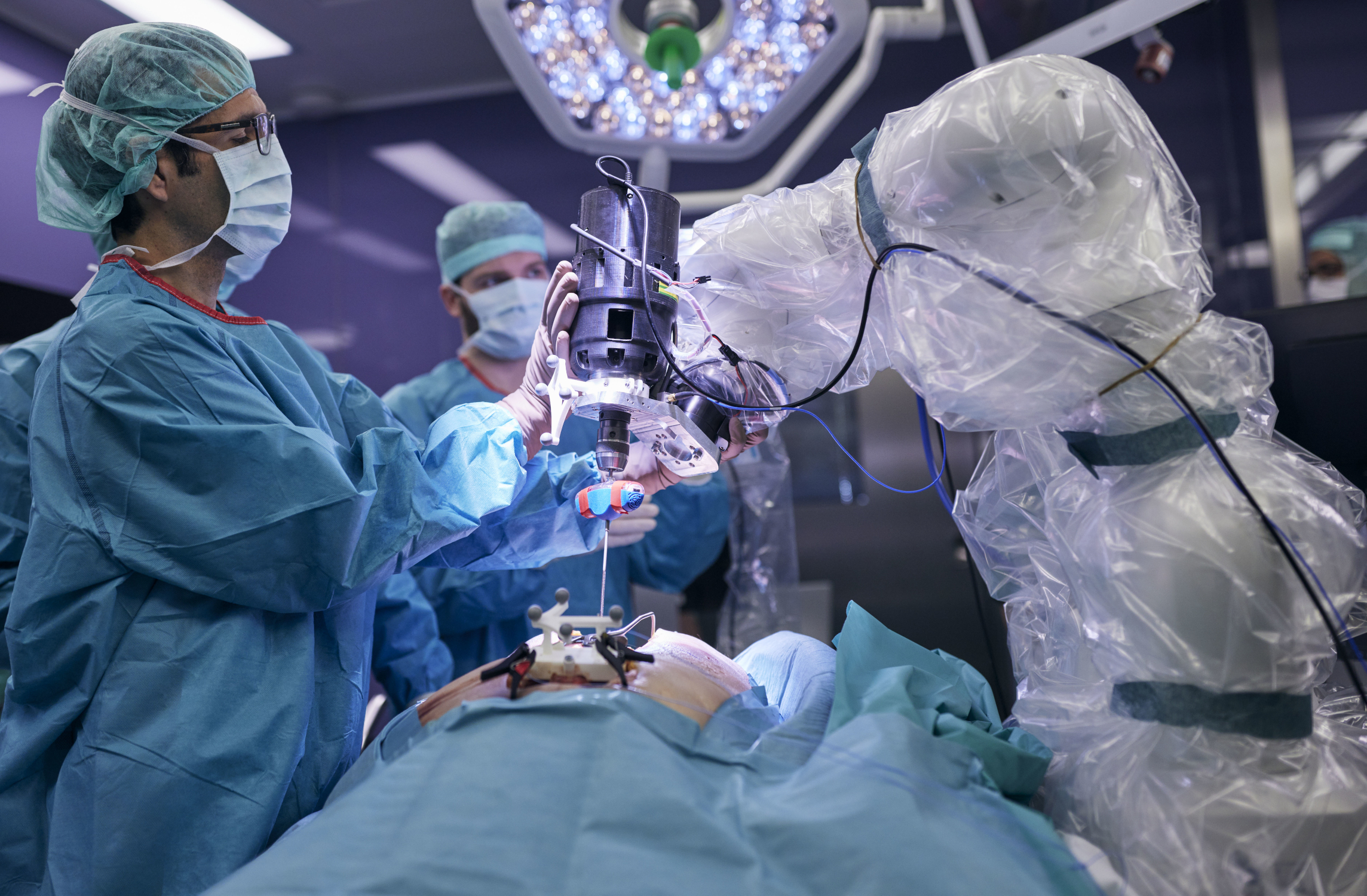}
    \caption{}
    \label{fig:suturing}
  \end{subfigure}
  \caption{Examples of contact-rich manipulation showing (a) a robot placing an item on a shelf, (b) a human interacting with a robot performing a drilling task during pedicle screw fixation. Image credit: University Hospital Balgrist, Daniel Hager Photography \& Film GmbH.
  }
  \label{fig:examples}
\end{figure}

High-dimensional motion planners and controllers are integrated in many of the approaches for solving complex manipulation tasks.
Consider, for example, a robot operating in an unstructured and dynamic environment that, e.g. places an object onto a shelf, or drilling during pedicle screw fixation in surgery (see Fig.\ \ref{fig:examples}).
In such cases, a planner and controller must account for objectives/constraints like bi-manual coordination, contact constraints between robot-object and object-environment, and be robust to disturbances.
Efficient motion planning and fast controllers are an effective way of enabling robots to perform these tasks subject to motion constraints, system dynamics, and changing task objectives.

Sampling-based planners~\cite{lavalle2006planning} are effective, however, they typically require considerable post-processing (e.g. trajectory smoothing).
Optimal planners (i.e. that are provably asymptotically optimal, e.g. RRT$^*$) are promising but inefficient (in terms of computation duration) for solving high-dimensional problems~\cite{karaman2011sampling}.

Gradient-based trajectory optimization (TO) is a key approach in optimal control, and has also been utilized for motion planning.
This approach underpins many recent works in robotics for planning and control, e.g.~\cite{Ratliff2009, Schulman2014, Posa2014, Kuindersma2016, stouraitis2020online, mower2021skill, moura2022non, Toussaint2022}.
Given an initialization, optimization finds a locally optimal trajectory, comprised of a stream of state and control commands subject to motion constraints and system dynamics (i.e. equations of motion).

Several reliable open-source and commercial optimization solvers exist for solving TO problems, e.g. IPOPT~\cite{Wachter2006}, KNITRO~\cite{byrd2006k}, and SNOPT~\cite{Gill2005SNOPT}.
However, despite the success of the optimization approaches proposed in the literature and motion planning frameworks such as MoveIt~\cite{coleman2014reducing},
there is a lack of libraries enabling fast development/prototyping of optimization-based approaches for multi-robot setups that easily interfaces with these efficient solvers.

To fill this gap, this paper proposes OpTaS, a user-friendly task-specification library for rapid development and deployment of nonlinear optimization-based planning and control approaches such as Model Predictive Control (MPC).
The library leverages the symbolic framework of CasADi \cite{Andersson2019}, enabling function derivatives to arbitrary order via automatic differentiation.
This is important since some solvers (e.g. SNOPT) utilize the Jacobian and Hessian. 

\begin{figure*}[t]
  \centering
  \vspace{0.65cm}
  \includegraphics[width=1.8\columnwidth]{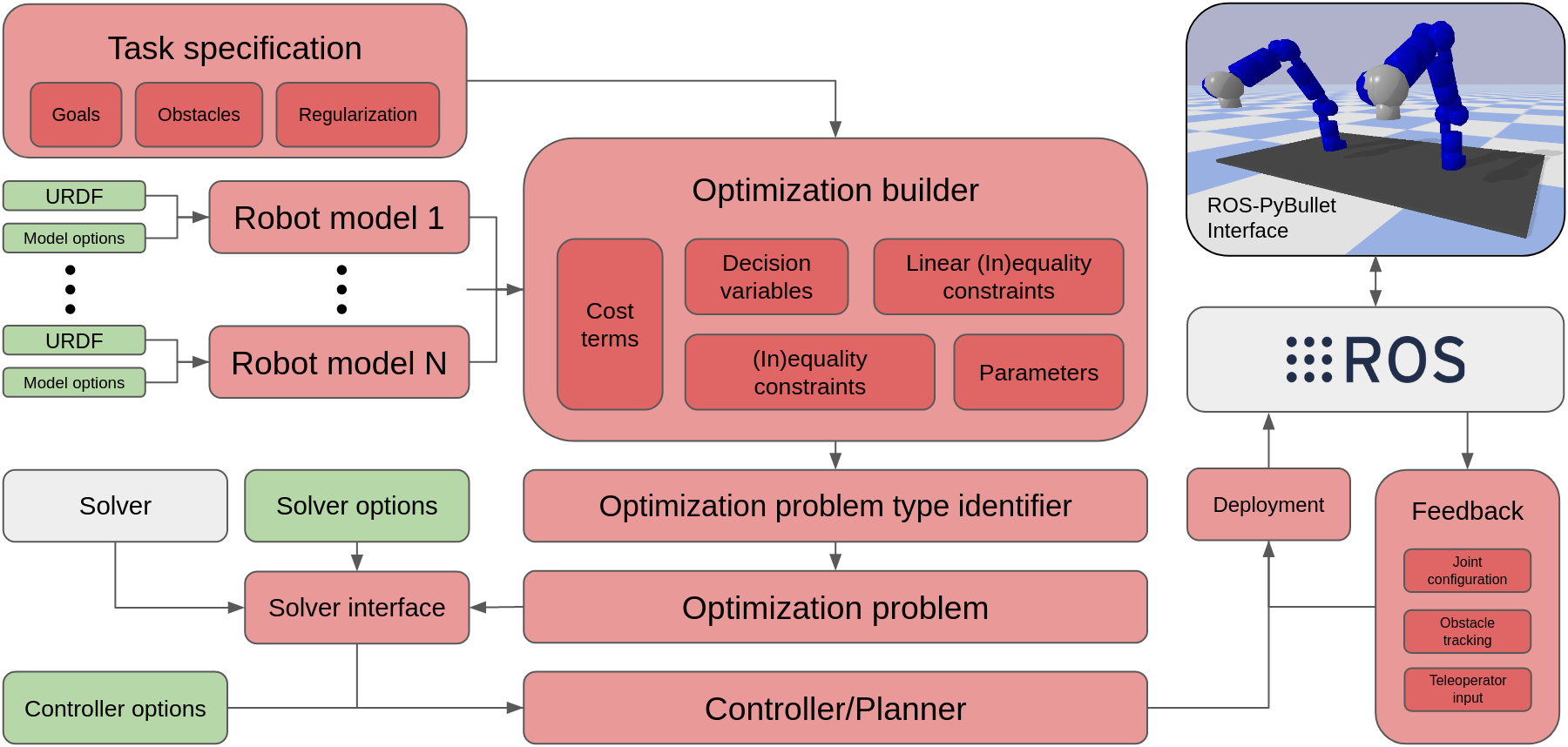}
  \caption{System overview for the proposed OpTaS library.
    \textbf{Red} highlights the main features of the proposed library.
    \textbf{Green} shows configuration parameter input.
    \textbf{Grey} shows third-party frameworks/libraries.
    Finally, the image in the top-right corner shows integration with the  ROS-PyBullet Interface~\cite{Mower2022}.
  }
  \label{fig:sys-overview}
   \vspace{-0.75cm}
\end{figure*}


\subsection{Related work}\label{sec:related-work}

In this section, we review popular optimization solvers and their interfaces.
Next, we describe works similar (in formulation) to our proposed library.
Finally, we summarize the key differences and highlight our contributions.
Table~\ref{tab:compare} summarizes alternatives and how they compare to OpTaS.

There are several capable open-source and commercial optimization solvers.
First considering quadratic programming, the OSQP method provides a general purpose solver based on the alternating direction method of multipliers~\cite{osqp}.
Alternatively, CVXOPT implements a custom interior-point solver~\cite{andersen2020cvxopt}.
IPOPT implements an interior-point solver for constrained nonlinear optimization.
SNOPT provides an interface to an SQP algorithm~\cite{Gill2005SNOPT}.
KNITRO also solves general mixed-integer programs~\cite{byrd2006k}.
Please note that SNOPT and KNITRO are proprietary.

These solvers are often implemented in low-level programming languages such as C, C++, or FORTRAN.
However, there are also many interfaces to these methods via higher level languages, such as Python, to make implementation and adoption easier.
The SciPy library contains the \texttt{optimize} module \cite{2020SciPy-NMeth} to interface with low-level routines, e.g. conjugate gradient and BFGS algorithm \cite{nocedal1999numerical}, the Simplex method \cite{nelder1965simplex}, COBYLA \cite{Powell1994}, and SLSQP \cite{Kraft1988}.
A requirement when using optimization-based methods is the need for function gradients.
Several popular software packages implement automatic differentiation~\cite{jax2018github, Andersson2019, NEURIPS2019_9015}.
We leverage the CasADi framework \cite{Andersson2019} for deriving gradients.
Our choice for CasADI is based on the fact that it comes readily integrated with common solvers for optimal control.
To the best of our knowledge, JAX and PyTorch are not currently integrated with constrained nonlinear optimization solvers.

Similar to our proposed library are the following packages.
The MoveIt package provides the user with specific IK/planning formulations and provides interfaces to solvers for the particular problem~\cite{coleman2014reducing}.
The eTaSL library \cite{Aertbelien2014etasl}  allows the user to specify custom tasks specifications, but only supports problems formulated as quadratic programs.
The CASCLIK library uses CasADi and provides support for constraint-based inverse kinematic controllers~\cite{Arbo2019}, to the best of our knowledge they allow optimization in the joint space. 
We provide joint space, task space optimization and also the ability to simultaneously optimize in the joint/task space. 
Furthermore, our framework supports optimization of several robots in a single formulation.
The EXOTica library allows the user to specify a problem formulation from an XML file \cite{exotica}. The package, however, requires the user to supply analytic gradients for additional sub-task models.

\begin{table}[]
\setlength{\tabcolsep}{2pt}
\def\t{\ding{51}} 
\def\c{\ding{55}} 
\scriptsize
\centering
\caption{Comparison between OpTaS and common alternatives in literature.}\label{tab:compare}
\begin{tabular}{l|llllllll}
               & Languages  & End-pose & Traj. & MPC & Solver & AutoDiff & ROS & Re-form \\ \hline
\textbf{OpTaS} & Python     & \t        & \t          & \t   & QP/NLP         & \t         & \t   & \t              \\
EXOTica        & Python/C++ & \t        & \t          & \c   & QP/NLP         & \c        & \t   & \t              \\
MoveIt         & Python/C++ & \t        & \t          & \c   & QP             & \c        & \t   & \c              \\
TracIK         & Python/C++ & \t        & \c          & \c   & QP             & \c        & \t   & \c             \\
RBDL           & Python/C++ & \t        & \c          & \c   & QP             & \c         & \c   & \c          \\
eTaSL         & C++        & \t        & \c          & \c   & QP             & \t         & \c\tablefootnote{Enabled with external pluggins.}  & \t              \\
OpenRAVE       & Python     & \c        & \t          & \c   & QP             & \c         & \t   & \c         
\end{tabular}
\end{table}

\subsection{Contributions}

This paper makes the following contributions:
\begin{itemize}
\item A task-specification library, in Python, for rapid development/deployment of TO approaches for multi-robot setups.
\item Modeling of the robot kinematics (forward kinematics, geometric Jacobian, etc.), to arbitrary derivative order, given a URDF specification.
\item An interface that allows a user to easily reformulate an optimal control problem, and define parameterized constraints for online modification of the optimization problem. 
\item Analysis comparing the performance of the library (i.e. solver convergence, solution quality) versus existing software packages. Further demonstrations highlight the ease in which nonlinear constrained optimization problems can be set up and deployed in realistic settings.
\end{itemize}

\section{Problem Formulation}\label{sec:problem-formulation}

We can write an optimal control formulation of a TO or planning problems as
\begin{equation}
  \label{eq:trajopt}
  \underset{x, u}{\min}~\text{cost}(x, u; T)\quad\text{subject to}\quad
  \begin{cases}
    \dot{x} = f(x, u)\\x\in\mathbb{X}\\u\in\mathbb{U}
  \end{cases}
\end{equation}
where $t$ denotes time, and $x = x(t)\in\mathbb{R}^{n_x}$ and $u = u(t)\in\mathbb{R}^{n_u}$ denote the states and controls, with $T$ being the time-horizon for the planned trajectory.
The scalar function $\text{cost}: \mathbb{R}^{n_x}\times \mathbb{R}^{n_u}\rightarrow\mathbb{R}$ represents the cost function (typically a weighted sum of terms each modeling a certain sub-task),
the dot notation denotes a derivative with respect to time (i.e. $\dot{x}\equiv\tfrac{dx}{dt}$),
$f$ represents the system dynamics (equations of motion), and
$\mathbb{X}\subseteq\mathbb{R}^{n_x}$ and $\mathbb{U}\subseteq\mathbb{R}^{n_u}$ are feasible regions for the states and controls respectively (modeled by a set of equality and inequality constraints).
Direct optimal control, optimizes for the controls $u$ for a discrete set of time instances, using numerical methods (e.g. Euler or Runge-Kutta), to integrate the system dynamics over the time horizon $T$~\cite{kelly2017introduction}.
Given an initialization $x^{\textrm{init}}, u^{\textrm{init}}$, a locally optimal trajectory $x^*, u^*$ is found by solving \eqref{eq:trajopt}.

As discussed in Sec.\ \ref{sec:intro}, many works propose optimization-based approaches for planning and control.
These can all be formulated under the same framework, i.e. a TO problem as in \eqref{eq:trajopt}.
The goal of our work is to deliver a library that allows a user to quickly develop and prototype constrained nonlinear TO for multi-robot problems, and deploy them for motion generation.
The library includes two types of problems, IK and task-sace TO, and indeed both simultaneously.
Common steps, such as transcription that transforms the problem's task-level description into a form accepted by numerical optimization solver routines, should be automated and thus not burden the user.
Furthermore, many works in practice require the ability to adapt constraints dynamically to handle changes in the environment (e.g. MPC).
This motivates a constraint parameterization feature.

\section{Proposed Framework}\label{sec:proposed-framework}

In this section, we describe the main features of the proposed library shown in Fig.\ \ref{fig:sys-overview}.
The library is completely implemented in the Python programming language.
We chose Python because it is simple for beginners but also versatile with many well-developed libraries, and it easily facilitates fast prototyping.

\subsection{Robot model}\label{sec:robot-model}

The robot model (\texttt{RobotModel}) provides the kinematic modeling and specifies the time derivative orders required for the optimization problem.
The only requirement is a URDF to instantiate the object\footnote{\href{http://wiki.ros.org/urdf}{http://wiki.ros.org/urdf}}.
A key feature is that we can include several robots in the TO, which is useful for dual arm and whole-body optimization.
Additional base frames and end-effector links can be added programatically (for example, when several robots are included the optimization their base frames should be registered within a global coordinate frame).

The \texttt{RobotModel} class allows access to data such as: the number of degrees of freedom, the names of the actuated joints, the upper and lower actuated joint limits, and the kinematics model.
Furthermore, we provide methods to compute the forward kinematics and geometric Jacobian in any given reference frame.
Several methods modeling the kinematics are supplied, given a specification from the user for the base frame and end-effector frame.
These methods include:
the $4\times 4$ homogeneous transformation matrix,
translation position,
rotational representations (e.g. Euler angles, quaternions),
the geometric and analytical Jacobian.
Each of the methods above depend on a joint state (supplied as either a Python list, NumPy array, or CasADi symbolic array).

\subsection{Task model}

Several works optimize robot motion in the task space and then compute the IK as a secondary step, e.g.~\cite{mower2021skill, moura2022non}.
The task model (\texttt{TaskModel}) provides a representation for any arbitrary trajectory. 
For example, the three dimensional position trajectory of an end-effector.
In the same way as the robot model, the time derivatives can be specified in the interface an arbitrary order.

\subsection{Optimization builder}

This section introduces and describes the optimization builder class (\texttt{OptimizationBuilder}).
The purpose of this class is to aid the user to easily setup a TO problem, and then automatically build an optimization problem model (Sec.\ \ref{sec:optimization-problem-model}) that interfaces with a solver interface (Sec.\ \ref{sec:solver-interface}).
The development cycle consists in specifying the task (i.e. decision variables, parameters, cost function, and constraints) using intuitive syntax and symbolic variables.
Then, the builder creates an optimization problem class, which interfaces with several solvers.

\subsection{Optimization problem model}\label{sec:optimization-problem-model}

The standard TO is stated in \eqref{eq:trajopt}.
This task/problem is specified by the optimization builder class in intuitive syntax for the user.
Transcribing the problem to a form that can be solved by off-the-shelf solvers is non-trivial.
The output of the optimization builder method \texttt{build} is an optimization problem model that allows us to interface with several solvers.

The most general optimization problem that is modeled by OpTaS is given by
\begin{subequations}
  \label{eq:optimization-problem}
  \begin{align}
    X^* & = \underset{X}{\text{arg}\min}~f(X; P)\\
        & \text{subject to}\nonumber\\
        & k(X; P) = M(P)X + c(P) \geq 0\label{eq:lin-ineq-con}\\
        & a(X; P) = A(P)X + b(P) = 0\label{eq:lin-eq-con}\\
        & g(X; P) \geq 0\label{eq:nlin-ineq-con}\\
        & h(X; P) = 0\label{eq:nlin-eq-con}
  \end{align}
\end{subequations}
where
$X = [vec(x)^T, vec(u)^T]^T\in\mathbb{R}^{n_X}$ is the decision variable array such that $x, u$ are as defined in \eqref{eq:trajopt} and $vec(\cdot)$ is a function that returns its input as a 1-dimensional vector,
$P\in\mathbb{R}^{n_P}$ is the vectorized parameters,
$f: \mathbb{R}^{n_X}\rightarrow\mathbb{R}$ denotes the objective function,
$k: \mathbb{R}^{n_X}\rightarrow\mathbb{R}^{n_k}$ denotes the linear inequality constraints,
$a: \mathbb{R}^{n_X}\rightarrow\mathbb{R}^{n_a}$ denotes the linear equality constraints,
$g: \mathbb{R}^{n_X}\rightarrow\mathbb{R}^{n_g}$ denotes the nonlinear inequality constraints, and
$h: \mathbb{R}^{n_X}\rightarrow\mathbb{R}^{n_h}$ denotes the nonlinear equality constraints.
The decision variables $X$ are all the joint states and other variables specified by the user stacked into a single vector.
Similarly for the parameters, cost terms, and constraints.
Vectorization is made possible by the \texttt{SXContainer} data structure implemented in the \texttt{sx\_container} module.
This data structure enables automatic transcription of the TO problem specified in \eqref{eq:trajopt} into the form \eqref{eq:optimization-problem}.

Of course, not all task specifications will require definitions for each of the functions in \eqref{eq:optimization-problem}.
Depending on the structure of the objective function and constraints, the required time budget, and accuracy, some solvers will be more appropriate for solving \eqref{eq:optimization-problem}.
For example, a quadratic programming solver that only handles linear constraints (e.g. OSQP \cite{osqp}) is unsuitable for solving a problem with nonlinear objective function and nonlinear constraints.
The build process automatically identifies the optimization problem type, exposing only the relevant solvers.
Several problem types are available to the user: 
unconstrained quadratic cost, 
linearly constrained with quadratic cost, 
nonlinear constrained with quadratic cost, 
unconstrained with nonlinear cost, 
linearly constrained with nonlinear cost,
nonlinear cost and constraints.

\subsubsection{Initialization}

\begin{figure}[t]
  \centering
  \vspace{0.25cm}
  \includegraphics[width=0.8\columnwidth]{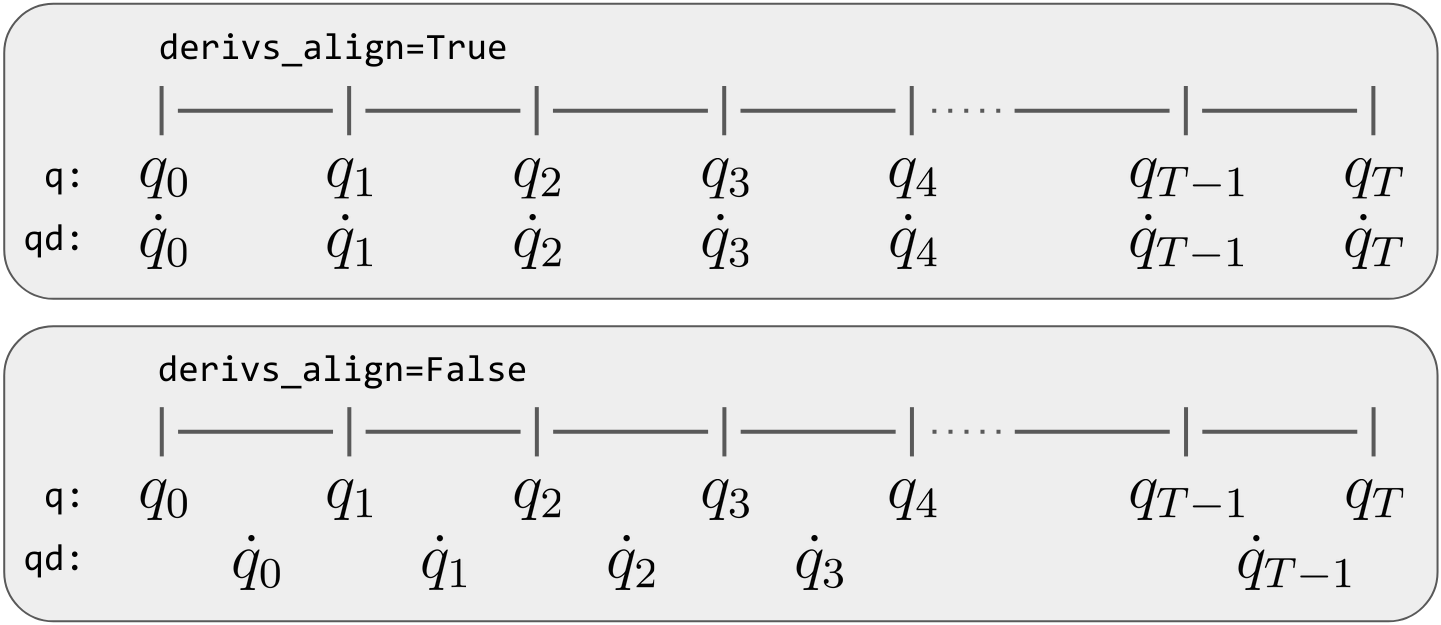}
  \caption{Joint state alignment with time. User supplies \texttt{derivs\_align} that specifies how joint state time derivatives should be aligned.}
  \label{fig:time-deriv}
  \vspace{-0.75cm}
\end{figure}

Upon initialization of the optimization builder class we can specify
\textbf{(i)} the number of time steps in the trajectory,
\textbf{(ii)} several robot and task models (given a unique name for each),
\textbf{(iii)} the joint states (positions and required time-derivatives) that integrate the decision variable array,
\textbf{(iv)} task space labels, dimensions, and derivatives to also integrate the decision variable array,
\textbf{(v)} a Boolean describing the alignment of the derivatives (Fig.\ \ref{fig:time-deriv}), and
\textbf{(vi)} a Boolean indicating whether to optimize time steps.

The alignment of time-derivatives can be specified in two ways.
Each derivative is aligned with its corresponding state (alignement), or otherwise.
This is specified by the \texttt{derivs\_align} flag in the optimization builder interface and shown diagramatically in Fig.\ \ref{fig:time-deriv}.


In addition, the user can also optimize the time-steps between each state.
The time derivatives can be integrated over time, e.g. $q_{t+1} = q_t + \delta\tau_t\dot{q}_t$, where $\delta\tau_t$ is an increment in time.
When \texttt{optimize\_time=True}, then each $\delta\tau_t$ is included as decision variables in the optimal control problem.

\subsubsection{Decision variables and parameters}

Decision variables are specified in the optimization builder class interface for the joint space, task space, and time steps.
Each group of variables is given a unique label and can be retrieved using the \texttt{get\_model\_state} method.
States are retrieved by specifying a robot name or task name, the required time index, and the time derivative order required.
Additional decision variables can be included in the problem by using the \texttt{add\_decision\_variables} method given a unique name and dimension.

Parameters for the problem (e.g. safe distances) can be specified using the \texttt{add\_parameter} method.
To specify a new parameter, a unique name and dimension is required.

\subsubsection{Cost and constraint functions}

The cost function in \eqref{eq:trajopt} is assumed to be made up of several cost terms, i.e.
\begin{equation}
  \label{eq:cost-function}
  \text{cost}(x, u; T) = \sum_{i}~c_i(x, u; T)
\end{equation}
where $c_i: \mathbb{R}^{n_x}\times\mathbb{R}^{n_u}\rightarrow\mathbb{R}$ is an individual cost term modeling a specific sub-task.
For example, let us define the cost terms $c_0 = \|\psi(x_T) - \psi^*\|^2$ and $c_1 = \lambda\int_0^T~\|u\|^2~dt$ (note, discretization is implicit in this formulation) where
$\psi:\mathbb{R}^{n_x}\rightarrow\mathbb{R}^3$ is a function for the forward kinematics position (note, this can be provided by the robot model class as described in Sec.\ \ref{sec:robot-model}),
$\psi^*\in\mathbb{R}^3$ is a goal task space position, and
$0<\lambda\in\mathbb{R}$ is a scaling term used to weight the relative importance of one constraint against the other.
Thus, $c_0$ describes an ideal state for the final state, and $c_1$ encourages trajectories with minimal control signals (e.g. minimize joint velocities).
Each cost term is added to the problem using the \texttt{add\_cost\_term} method; the \texttt{build} sequence ensures each term is added to the objective function.

Several constraints can be added to the optimization problem by using the \texttt{add\_equality\_constraint} and \texttt{add\_leq\_inequality\_constraint} methods that add equality and inequality constraints respectively.
When the constraints are added to the problem, they are first checked to see if they are linear constraints with respect to the decision variables.
This functionality allows the library to differentiate between linear and nonlinear constraints.

Additionally, OpTaS offers several methods that provide an implementation for common constraints, as, for example, joint position/velocity limits and time-integration for the system dynamics $f$ (e.g joint velocities can be integrated to positions).

\subsection{Solver interface}\label{sec:solver-interface}

OpTaS provides interfaces to solvers (open-source and commercial) that interface with CasADi~\cite{Andersson2019} (such as IPOPT~\cite{Wachter2006}), SNOPT~\cite{Gill2005SNOPT}, KNITRO~\cite{byrd2006k}, and Gurobi~\cite{gurobi}), the Scipy \texttt{minimize} method~\cite{2020SciPy-NMeth}, OSQP~\cite{osqp}, and CVXOPT~\cite{andersen2020cvxopt}.

\subsubsection{Initialization of solver}

When the solver is initialized, several variables are setup and the optimization problem object is set as a class attribute.
The user must then call the \texttt{setup} method - that itself is an interface to the solver initialization that the user has chosen.
The requirement of this method is to setup the interface for the specific solver; relevant solver parameters are passed to the interface at this stage.

\subsubsection{Resetting the interface}

When using the solver as a controller, it is expected that the solver should be called more than once.
In the case for feedback controllers or controllers with parameterized constraints (e.g. obstacles), this requires a way to reset the problem parameters.
Furthermore, the initial seed for the optimizer is often required to be reset at each control loop cycle.
To reset the initial seed and problem parameters the user calls \texttt{reset\_initial\_seed}, and \texttt{reset\_parameters}, respectively.
Both the initial seed and parameters are initialized by giving the name of the variables. The required vectorization is internally performed by the solver utilizing features of the \texttt{SXContainer} data structure.
Note, if any decision variables or parameters are not specified in the reset methods then they automatically default to zero.
This enables warm-starting the optimization routine, e.g. with the solution of the previous time-step problem.

\subsubsection{Solving an optimization problem}

The optimization problem is solved by calling the \texttt{solve} method.
This method passes the optimization problem to the desired solver.
The resulting data from the solver is collected and transformed back into the state trajectory for each robot.
A method is provided, named \texttt{interpolate}, is used to interpolate the computed trajectories across time.
Additionally, the method \texttt{stats} retrieves available optimization statistics (e.g. number of iterations).

\subsubsection{Extensible solver interface}

The solver interface has been implemented to allow for extensibility, i.e.  additional optimization solvers can be easily integrated into the framework.
When a user would like to include a new solver interface, they must create a new class that inherits from the \texttt{Solver} class.
In their sub-class definition they must implement three methods:
(i) \texttt{setup} which (as described above) initializes the solver interface,
(ii) \texttt{\_solve} that calls the solver and returns the optimized variable $X^*$, and
(iii) \texttt{stats} that returns any statistics from the solver.

\begin{figure}[t]
  \centering
   \vspace{0.25cm}


    



\begin{Verbatim}[fontsize=\footnotesize, commandchars=\\\{\}]
    \PYG{k+kn}{import} \PYG{n+nn}{optas}

    \PYG{c+c1}{\PYGZsh{} Setup robot and optimization builder}
    \PYG{n}{T} \PYG{o}{=} \PYG{l+m+mi}{100} \PYG{c+c1}{\PYGZsh{} number of time steps in trajectory}
    \PYG{n}{urdf} \PYG{o}{=} \PYG{l+s+s1}{\PYGZsq{}/path/to/robot.urdf\PYGZsq{}}
    \PYG{n}{r} \PYG{o}{=} \PYG{n}{optas}\PYG{o}{.}\PYG{n}{RobotModel}\PYG{p}{(}\PYG{n}{urdf}\PYG{p}{,} \PYG{n}{time\PYGZus{}deriv}\PYG{o}{=}\PYG{p}{[}\PYG{l+m+mi}{0}\PYG{p}{,} \PYG{l+m+mi}{1}\PYG{p}{])}
    \PYG{n}{n} \PYG{o}{=} \PYG{n}{r}\PYG{o}{.}\PYG{n}{get\PYGZus{}name}\PYG{p}{()}
    \PYG{n}{b} \PYG{o}{=} \PYG{n}{optas}\PYG{o}{.}\PYG{n}{OptimizationBuilder}\PYG{p}{(}\PYG{n}{T}\PYG{o}{=}\PYG{n}{T}\PYG{p}{,} \PYG{n}{robots}\PYG{o}{=}\PYG{p}{[}\PYG{n}{r}\PYG{p}{])}

    \PYG{c+c1}{\PYGZsh{} Retrieve variables and setup parameters}
    \PYG{n}{q0} \PYG{o}{=} \PYG{n}{b}\PYG{o}{.}\PYG{n}{get\PYGZus{}model\PYGZus{}state}\PYG{p}{(}\PYG{n}{n}\PYG{p}{,} \PYG{n}{t}\PYG{o}{=}\PYG{l+m+mi}{0}\PYG{p}{)}
    \PYG{n}{qT} \PYG{o}{=} \PYG{n}{b}\PYG{o}{.}\PYG{n}{get\PYGZus{}model\PYGZus{}state}\PYG{p}{(}\PYG{n}{n}\PYG{p}{,} \PYG{n}{t}\PYG{o}{=\PYGZhy{}}\PYG{l+m+mi}{1}\PYG{p}{)} \PYG{c+c1}{\PYGZsh{} final state}
    \PYG{n}{pg} \PYG{o}{=} \PYG{n}{b}\PYG{o}{.}\PYG{n}{add\PYGZus{}parameter}\PYG{p}{(}\PYG{l+s+s1}{\PYGZsq{}pg\PYGZsq{}}\PYG{p}{,} \PYG{l+m+mi}{3}\PYG{p}{)} \PYG{c+c1}{\PYGZsh{} goal pos.}
    \PYG{n}{qc} \PYG{o}{=} \PYG{n}{b}\PYG{o}{.}\PYG{n}{add\PYGZus{}parameter}\PYG{p}{(}\PYG{l+s+s1}{\PYGZsq{}qc\PYGZsq{}}\PYG{p}{,} \PYG{n}{r}\PYG{o}{.}\PYG{n}{ndof}\PYG{p}{)} \PYG{c+c1}{\PYGZsh{} init q}
    \PYG{n}{o} \PYG{o}{=} \PYG{n}{b}\PYG{o}{.}\PYG{n}{add\PYGZus{}parameter}\PYG{p}{(}\PYG{l+s+s1}{\PYGZsq{}o\PYGZsq{}}\PYG{p}{,} \PYG{l+m+mi}{3}\PYG{p}{)} \PYG{c+c1}{\PYGZsh{} obstacle pos.}
    \PYG{n}{r} \PYG{o}{=} \PYG{n}{b}\PYG{o}{.}\PYG{n}{add\PYGZus{}parameter}\PYG{p}{(}\PYG{l+s+s1}{\PYGZsq{}r\PYGZsq{}}\PYG{p}{)}  \PYG{c+c1}{\PYGZsh{} obstacle radius}
    \PYG{n}{dt} \PYG{o}{=} \PYG{n}{b}\PYG{o}{.}\PYG{n}{add\PYGZus{}parameter}\PYG{p}{(}\PYG{l+s+s1}{\PYGZsq{}dt\PYGZsq{}}\PYG{p}{)} \PYG{c+c1}{\PYGZsh{} time step}

    \PYG{c+c1}{\PYGZsh{} Forward kinematics}
    \PYG{n}{p} \PYG{o}{=} \PYG{n}{r}\PYG{o}{.}\PYG{n}{get\PYGZus{}global\PYGZus{}link\PYGZus{}position}\PYG{p}{(}\PYG{n}{tip}\PYG{p}{,} \PYG{n}{qT}\PYG{p}{)}

    \PYG{c+c1}{\PYGZsh{} Cost and constraints}
    \PYG{n}{b}\PYG{o}{.}\PYG{n}{add\PYGZus{}cost\PYGZus{}term}\PYG{p}{(}\PYG{l+s+s1}{\PYGZsq{}c\PYGZsq{}}\PYG{p}{,} \PYG{n}{optas}\PYG{o}{.}\PYG{n}{sumsqr}\PYG{p}{(}\PYG{n}{p} \PYG{o}{\PYGZhy{}} \PYG{n}{pg}\PYG{p}{))}
    \PYG{n}{b}\PYG{o}{.}\PYG{n}{integrate\PYGZus{}model\PYGZus{}states}\PYG{p}{(}
        \PYG{n}{n}\PYG{p}{,} \PYG{n}{time\PYGZus{}deriv}\PYG{o}{=}\PYG{l+m+mi}{1}\PYG{p}{,} \PYG{n}{dt}\PYG{o}{=}\PYG{n}{dt}\PYG{p}{)}
    \PYG{n}{b}\PYG{o}{.}\PYG{n}{add\PYGZus{}equality\PYGZus{}constraint}\PYG{p}{(}\PYG{l+s+s1}{\PYGZsq{}init\PYGZsq{}}\PYG{p}{,} \PYG{n}{q0}\PYG{p}{,} \PYG{n}{qc}\PYG{p}{)}
    \PYG{k}{for} \PYG{n}{t} \PYG{o+ow}{in} \PYG{n+nb}{range}\PYG{p}{(}\PYG{n}{T}\PYG{p}{):}
        \PYG{n}{b}\PYG{o}{.}\PYG{n}{add\PYGZus{}leq\PYGZus{}inequality\PYGZus{}constraint}\PYG{p}{(}
            \PYG{n}{optas}\PYG{o}{.}\PYG{n}{sumsqr}\PYG{p}{(}\PYG{n}{p} \PYG{o}{\PYGZhy{}} \PYG{n}{o}\PYG{p}{),} \PYG{n}{r}\PYG{o}{**}\PYG{l+m+mi}{2}\PYG{p}{)}

    \PYG{c+c1}{\PYGZsh{} Build optimization problem and setup solver}
    \PYG{n}{solver} \PYG{o}{=} \PYG{n}{optas}\PYG{o}{.}\PYG{n}{CasADiSolver}\PYG{p}{(}
        \PYG{n}{b}\PYG{o}{.}\PYG{n}{build}\PYG{p}{())}\PYG{o}{.}\PYG{n}{setup}\PYG{p}{(}\PYG{l+s+s1}{\PYGZsq{}ipopt\PYGZsq{}}\PYG{p}{)}
\end{Verbatim}

  \caption{Example code for TO described in Section \ref{sec:code-example}. 
  }
  \label{fig:code-example}
   \vspace{-0.75cm}
\end{figure}




\subsection{Additional features}


Support for integration with ROS~\cite{Quigley09} is provided out-of-the-box.
The ROS node provided is integrated with the ROS-PyBullet Interface \cite{Mower2022} so the publishers/subscribers can connect a robot in the optimization problem with a robot simulated in PyBullet.


In addition, we provide a port of the \texttt{spatialmath} library by Corke~\cite{Corke17a} that supports CasADi variables.
This library defines methods for manipulating homogeneous transformation matrices, quaternions, Euler angles, etc. using CasADi symbolic variables.

\section{Code Example}\label{sec:code-example}

In this section, we describe a common TO problem and give the code that models the problem.
We aim to highlight how straightforward it is to setup a problem.

Consider a serial link manipulator, and goal to find a collision-free plan over time horizon $T$ to a goal end-effector position $p_g$ given a starting configuration $q_c$.
A single spherical collision is represented by a position $o$ and radius $r$.
The robot configuration $q_t$ represent states, and the velocities $\dot{q}_t$ are controls.

The cost function is given by $\|p(q_T) - p_g\|^2$ where $p$ is the position of the end-effector given by the forward kinematics.
We solve the problem by minimizing the cost function subject to the constraints: 
(i) initial configuration, $q_0 = q_c$, 
(ii) joint limits $q^-\leq q_t\leq q^+$, and
(iii) obstacle avoidance, $\|p(q_t) - o\|^2\geq r^2$.
The system dynamics is represented by several equality constraints $q_{t+1} = q_t + \delta t\dot{q}_t$ that can be specified by methods already in-built into OpTaS. 
The code for the TO problem above, is shown in Fig.\ \ref{fig:code-example}.

\section{Experiments}\label{sec:experiments}

\begin{figure}
    \centering
    \vspace{0.25cm}
    
        \begin{subfigure}[b]{\columnwidth}
    \centering
    \includegraphics[width=0.6\columnwidth]{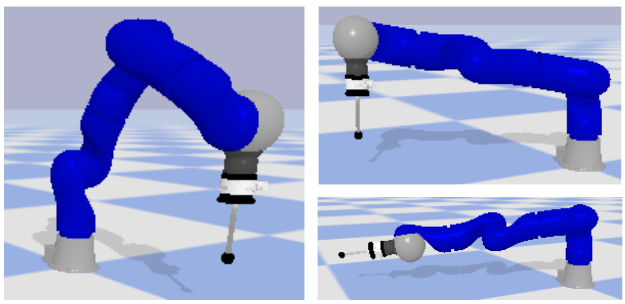}
    \caption{}
    \label{fig:kuka-start}
  \end{subfigure}\\
    
    \begin{subfigure}[b]{\columnwidth}
    \centering
    \includegraphics[width=0.9\columnwidth]{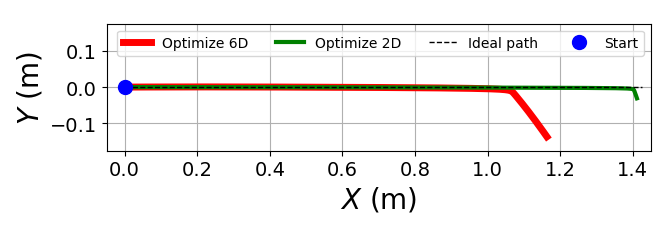}
    \caption{}
    \label{fig:plt-trajppos}
  \end{subfigure}
    
    \caption{Comparison of end-effector task space trajectories computed using two different formulations.
    (a) Shows the start (left), and final configurations (right) for the robot under each approach.
    (b) Plots the end-effector position trajectory two dimensions.}
    \label{fig:pos-traj}
    \vspace{-0.8cm}
\end{figure}


\subsection{Optimization along custom dimensions}

Popular solvers, such as TracIK~\cite{Beeson2015}, require the user to provide a 6D pose as the task space goal. 
Whilst this is applicable to several robotics problems (e.g. pick-and-place) it may not be necessary to optimize each task space dimension (e.g. spraying applications does not require optimization in the roll angular direction). 
Furthermore, optimizing in more dimensions than necessary may be disadvantageous.
 
OpTaS can optimize or neglect any desired task space dimension.
This can have certain advantages, for example increasing the robot workspace.
Consider a non-prehensile pushing task along the plane, optimizing the full 6D pose may not be ideal since the task is two dimensional.
By optimizing in the two dimensional plane and specifying boundary constraints on the third linear spatial dimension, increases the robots workspace.

We setup a tracking experiment in OpTaS using a simulated Kuka LWR robot arm to compare the two cases: (i) optimize the full 6D pose, and (ii) optimize 2D linear position.
The robot is given an initial configuration (Fig. \ref{fig:kuka-start} left) and the task is to move the end-effector with velocity of constant magnitude and direction in the 2D plane.
The end configuration for each approach is shown in Fig. \ref{fig:kuka-start} right and the end-effector trajectories are shown in Fig. \ref{fig:plt-trajppos}.
We see that the 2D optimization problem is able to reach a greater distance, highlighting that the robot workspace is increased.

\subsection{Performance comparison}

\begin{figure}
    \centering
    \vspace{0.25cm}
    \includegraphics[width=0.3\columnwidth]{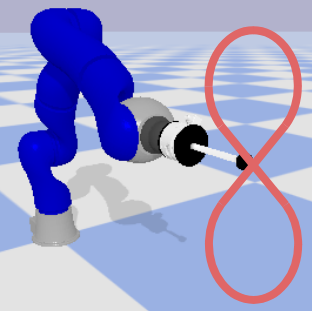}
    \caption{Figure-of-eight trajectory tracked by the Kuka LWR.}
    \label{fig:kuka-fig8}
\end{figure}

\begin{figure}[t]
  \centering

  \begin{subfigure}[b]{0.475\columnwidth}
    \centering
    
    \includegraphics[width=\columnwidth, trim={0.2cm 0.5cm 0.75cm 0.4cm}, clip]{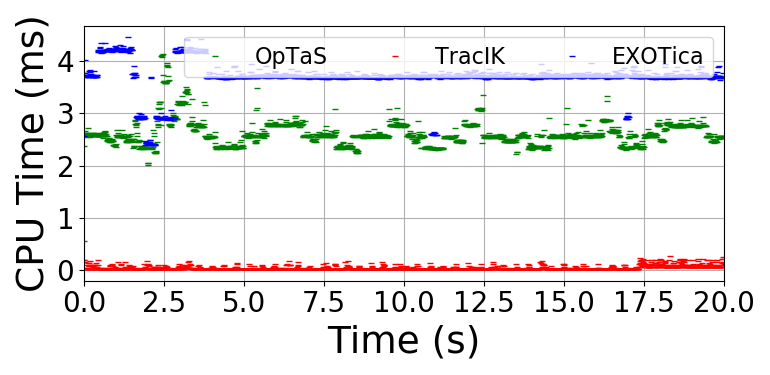}

    \caption{}
    \label{fig:time-comparison}
  \end{subfigure}
  ~
  \begin{subfigure}[b]{0.475\columnwidth}
    \centering
    \includegraphics[width=\columnwidth, trim={0.2cm 0.5cm 0.75cm 0.25cm}, clip]{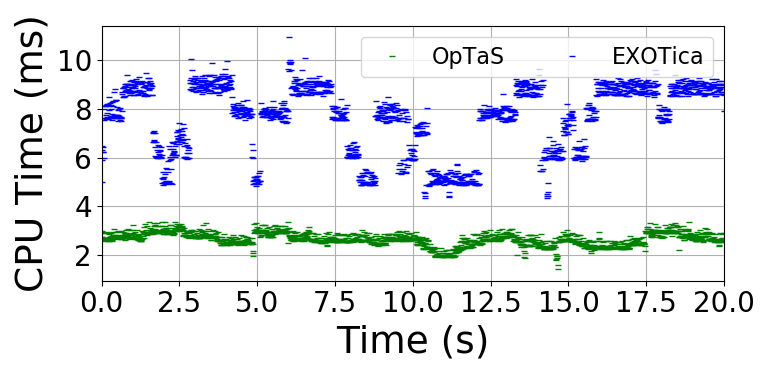}
    \caption{}
    
    \label{fig:err-comparison}
  \end{subfigure}
 \caption{Solver duration comparisons for figure of eight motion. (a) Compares an IK tracking approach described in Section \ref{sec:experiments}, (b) is a similar comparison that includes a maximization term for manipulability. Green is OpTaS, red is TracIK, and blue is EXOTica.}
   \label{fig:comparisons}
   \vspace{-0.75cm}
\end{figure}

In this section, we demonstrate that OpTaS can formulate similar problems and compare its performance to alternatives. 
First, we model, with OpTaS, the same problem as used in TracIK~\cite{Beeson2015} and in addition we also model the problem using EXOTica~\cite{exotica}.
The Scipy SLSQP solver~\cite{Kraft1988} was used for OpTaS and EXOTica.
With same Kuka LWR robot arm in the previous experiment, we setup a task where the robot must track a figure-of-eight motion in task space (Fig. \ref{fig:kuka-fig8}) and record the CPU time for the solver duration at each control loop cycle.
The results are shown in Fig. \ref{fig:time-comparison}.
TracIK is the fastest ($0.049\pm 0.035$ms), which is expected since it is optimized for a specific problem formulation.
We see that OpTaS ($2.608\pm 0.239$ms) is faster than EXOTica ($3.694\pm 0.300$ms)


A second experiment, using the same setup as before, was performed comparing the performance of OpTaS against EXOTica with an additional cost term to maximize manipulability~\cite{Yoshikawa85}.
The results are shown in Fig.\ \ref{fig:err-comparison}.
Despite using the same formulation and solver, OpTaS ($2.650\pm 0.270$ms) achieved better performance than EXOTica ($7.640\pm 1.404$ms).
Without extensive profiling it is difficult to precisely explain this difference.
However, EXOTica requires the user to supply analytical gradients for sub-tasks (called \textit{task maps} in the EXOTica documentation).
EXOTica does not provide the gradients for the manipulability task, and thus falls-back to using the finite difference method to estimate the gradient - this can can be slow to compute.




\section{Conclusions}\label{sec:conclusions}

In this paper, we have proposed OpTaS: an optimization-based task tpecification Python library for TO and MPC.
OpTaS allows a user to setup a constrained nonlinear programs for custom problem formulations and has been shown to perform well against alternatives.
Parameterization enables programs to act as feedback controllers, motion planners, and benchmark problem formulations and solvers.

We hope OpTaS will be used by researchers, students, and industry to facilitate the development of control and motion planning algorithms.
The code base is easily installed via \texttt{pip} and has been made open-source under the Apache 2 license:
\href{https://github.com/cmower/optas}{https://github.com/cmower/optas}.




%


\bibliographystyle{IEEETran}
\bibliography{bib}

\end{document}